\pdfoutput=1
\documentclass[11pt]{article}
\usepackage[]{acl}
\usepackage{times}
\usepackage{latexsym}
\usepackage[T1]{fontenc}
\usepackage[utf8]{inputenc}

\usepackage{microtype}

\usepackage{inconsolata}

\usepackage{listings}
\usepackage{multirow}
\usepackage{soul} 
\usepackage{booktabs}
\usepackage{graphicx}
\usepackage{subcaption}
\usepackage{graphics}
\usepackage{xspace}
\usepackage{amsmath}
\usepackage{amssymb}  
\usepackage{adjustbox}
\usepackage{makecell}
\usepackage{pifont}
\usepackage{tabularx}
\usepackage{xcolor}

\usepackage{tikz}
\usepackage{pgfplots}
\usepackage{xspace}
\usepackage{ulem}

\pgfplotsset{compat=1.18}

\definecolor{modelgreen}{RGB}{0,204,102}
\definecolor{modelred}{RGB}{255,0,0}
\definecolor{lightgray}{gray}{0.9}

\newcommand{\ourdata}{{\textsc{CustomPlans}}\xspace}
\newcommand{\easydata}{{\textsc{CustomPlansSimulated}}\xspace}
\newcommand{\diagdata}{{\textsc{CustomPlansReal}}\xspace}

\newcommand{\curie}{{\texttt{text-curie-001}}\xspace}

\newcommand{\turbo}{{\texttt{gpt-3.5-turbo}}\xspace}

\newcommand{\llm}{\textsc{LLM}}
\newcommand{\gptfour}{\textsc{GPT-4}\xspace}
\newcommand{\choice}{\textsc{Choice-75}\xspace}
\newcommand{\coplan}{\textsc{CoPlan}\xspace}

\newcommand{\cust}{\textsc{Customized}\xspace}
\newcommand{\exec}{\textsc{Executable}\xspace}
\newcommand{\fc}{\textsc{FullyCorrect}\xspace}

\newcommand{\custagent}{\texttt{Modify}\xspace agent~}
\newcommand{\execagent}{\texttt{Verify}\xspace agent~}

\newcommand{\paragents}{\textsc{Parallel}\xspace}
\newcommand{\seqagents}{\textsc{Sequential}\xspace}
\newcommand{\uniagent}{\textsc{Unified}\xspace}

\usepackage{soul}

\newcommand{\squishlist}{
  \begin{list}{$\bullet$}
    { \setlength{\itemsep}{0pt}      \setlength{\parsep}{3pt}
      \setlength{\topsep}{3pt}       \setlength{\partopsep}{0pt}
      \setlength{\leftmargin}{1.5em} \setlength{\labelwidth}{1em}
      \setlength{\labelsep}{0.5em} } }
\newcommand{\reallysquishlist}{
  \begin{list}{$\bullet$}
    { \setlength{\itemsep}{0pt}    \setlength{\parsep}{0pt}
      \setlength{\topsep}{0pt}     \setlength{\partopsep}{0pt}
      \setlength{\leftmargin}{0.2em} \setlength{\labelwidth}{0.2em}
      \setlength{\labelsep}{0.2em} } }

 \newcommand{\squishend}{
     \end{list} 
 }

\renewcommand{\cite}{\citep}

\definecolor{ipcolor}{RGB}{61, 133, 198}
\definecolor{modelcolor}{RGB}{241, 194, 50}
\definecolor{opcolor}{RGB}{106, 168, 79}

\definecolor{lightgray}{gray}{0.9}
\definecolor{Box1Color}{RGB}{227, 236, 246}
\definecolor{Box2Color}{RGB}{248, 220, 225}
\definecolor{Box3Color}{RGB}{255, 238, 224}
\definecolor{cbBlue}{RGB}{0, 114, 178}
\definecolor{cbOrange}{RGB}{240, 228, 66}
\definecolor{cbGreen}{RGB}{0, 158, 115}
\definecolor{cbRed}{RGB}{213, 94, 0}
\definecolor{cbPurple}{RGB}{204, 121, 167}
\definecolor{cbSkyBlue}{RGB}{86, 180, 233}
\definecolor{cbGray}{RGB}{128, 128, 128}
\definecolor{CBF1}{RGB}{255,99,132}  
\definecolor{CBF2}{RGB}{54,162,235}  
\definecolor{CBF3}{RGB}{255,206,86}  
\definecolor{CBF4}{RGB}{75,192,192}  
\definecolor{CBF5}{RGB}{153,102,255} 
\definecolor{CBF1b}{RGB}{205,89,112}  
\definecolor{CBF2b}{RGB}{44,142,215}  
\definecolor{CBF5b}{RGB}{133,92,225}  

\title{Tailoring with Targeted Precision: Edit-Based Agents for \newline Open-Domain Procedure Customization}

\author{Yash Kumar Lal$^1$$^3$\thanks{Work done as an intern at AI2 Aristo}, Li Zhang$^2$$^3$$^*$, Faeze Brahman$^3$, \\ \textbf{Bodhisattwa Prasad Majumder$^3$, Peter Clark$^3$, Niket Tandon$^3$} \\
\\
  $^1$ Stony Brook University,
  $^2$ University of Pennsylvania\\
  $^3$ Allen Institute for Artificial Intelligence \\
  $^1$\texttt{ylal@cs.stonybrook.edu},
  $^2$\texttt{zharry@upenn.edu},\\
  $^3$\texttt{\{faezeb,bodhisattwam,peterc,nikett\}@allenai.org}\\}

\begin{document}
\maketitle

\begin{abstract}
How-to procedures, such as how to plant a garden, are now used by millions
of users, but sometimes need customizing to meet a user's specific needs, e.g.,
planting a garden without pesticides. Our goal is to measure and improve an LLM's ability
to perform such customization. Our approach is to test several simple
multi-LLM-agent architectures for customization, as well as an end-to-end LLM,
using a new evaluation set, called \ourdata, of over 200 WikiHow procedures
each with a customization need. We find that a simple architecture with two LLM agents
used sequentially performs best, one that edits a generic how-to procedure
and one that verifies its executability, significantly outperforming 
(10.5\% absolute) an end-to-end prompted LLM. This suggests that LLMs can be
configured reasonably effectively for procedure customization. This also suggests
that multi-agent editing architectures may be worth exploring further for other customization
applications (e.g. coding, creative writing) in the future.
\end{abstract}

\section{Introduction}

AI is headed towards a future where human-machine interactions are seamlessly integrated to enrich our daily routines, offering personalized and tailored experiences  \cite{chen2023PersonalizationSurvey}. 
For instance, a software engineer's daily routine would involve a co-pilot that customizes the same underlying logic differently for two engineers (even though we observe the structure might broadly remain the same).
Another application is smart assistant and planners that can customize how-to procedures, a popular query making up a large fraction of search engine queries \cite{de2005question, zhang2022reasoning}, based on a user's specifications (though the sequence of steps broadly stay the same). 
For example, a user looking for ``How to plant a garden'', may have space restrictions in their apartment, or not want to use pesticides. 
Despite the need for customization, it is challenging to author new customized how-to procedures for every users' nuanced needs.

\begin{figure}[!t]
    \centering
    \includegraphics[width=\columnwidth]{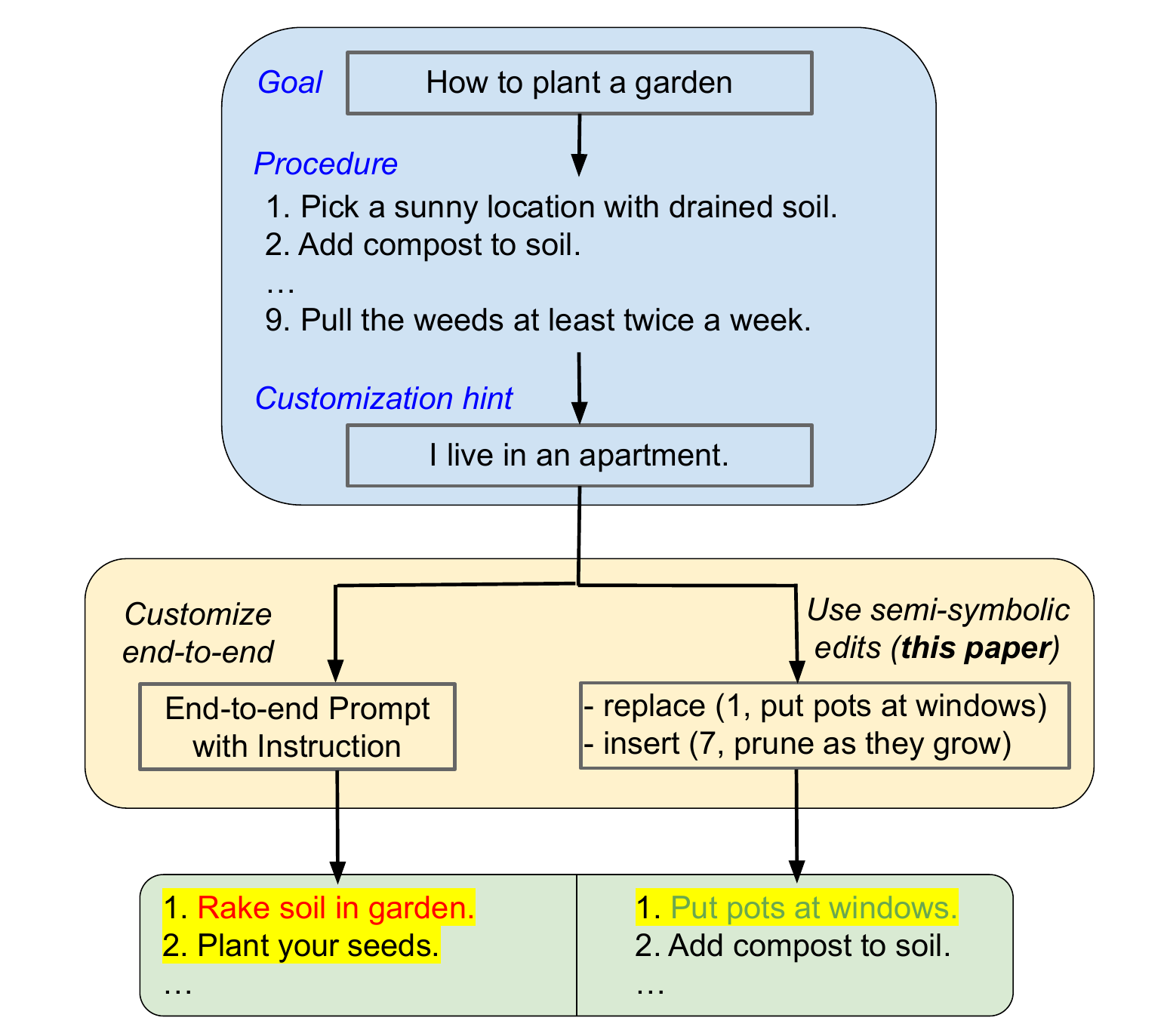}
    \caption{Procedures, e.g., how to plant a garden, need customization, e.g., this user lives in an apartment. Given a \textcolor{ipcolor}{goal}, \textcolor{ipcolor}{uncustomized procedure} and \textcolor{ipcolor}{customization hint}, employing \textcolor{modelcolor}{semi-symbolic edits} (right) produces more desirable \textcolor{opcolor}{outputs} than an E2E LLM (left; \textcolor{modelred}{which suggests a garden patch inside an apartment}).
}
    \label{fig:task_ex}
\end{figure}

Automatically customizing procedures requires the interpretation of the nuanced user needs expressed in natural language \cite{du2006understanding}. 
These customization requests or hints can take up various forms e.g., constraints (``I live in an apartment''), personal preferences (``I prefer organic farming''), or execution method (``plant a hydroponic garden"). 
These implicitly encode multiple requirements and their interpretation is subjective --- for example, living in an apartment can entail a lack of space, limited space, convincing roommates, and more.
Contemporary approaches to customization focus on constraints in specific domains \cite{yuan-etal-2023-distilling, welch-etal-2022-leveraging}. 
\llm s could be considered a strong baseline for faithfully customizing procedures to different users' needs \cite{acher-2024}, and we did find that the generated customized procedures are fluent and coherent. 
However, in our experiments \S\ref{subsec:e2e-llms}, we found that \textasciitilde60\% of the procedures generated by contained errors (missing steps, extra steps, wrong steps, underspecified steps), making the output inadequately customized or inexecutable, as shown in \autoref{fig:task_ex}.
We observe that even though uncustomized and customized procedures share some inherent structure, end-to-end systems disregard that and produce entirely new structures which introduces unwanted changes.

Rather than using \llm s as end-to-end customizers, we distenagle the task into modifying a procedure based on a customization requirement, and verifying for executability.
We propose a multi-agent framework comprising two \llm-based agents, \custagent and \execagent for customization and execution verification respectively. 
We create a new evaluation set called \ourdata of over 200 WikiHow procedures each with a customization need, and
show that these agents are most effective when operating based on semi-symbolic edits rather than free-form natural language edits. 
We also discuss the generalizability of our framework to support multiple \texttt{Verify} agents. 
Through extensive experiments with \ourdata, we find that our multi-agent framework leads to 10.5\% more customized and executable procedures over just using \llm s as end-to-end customizers.

In summary, our contributions are:
\squishlist
    \item Using a new evaluation set \ourdata, we show that \llm s are yet unsuited to customize how-to procedures in an end-to-end fashion.
    \item We propose a multi-agent framework comprising \texttt{Modify} and \texttt{Verify} agents, and show that semi-symbolic edits is the most effective means of communicating results. This framework achieves an improvement of 10.5\% over using \llm s as end-to-end customizers.
    \item We show the generalizability of our framework to support multiple agent configurations, and the limits of current methods when employed for broader applications.
\squishend

\section{Task Setup}
\label{sec:e2e-llms-eval}

In this section, we define a problem formulation and evaluation scheme for procedure customization and showcase the shortcomings of using \llm s as end-to-end customizers.
\llm s are capable of generating fluent texts, including procedures \cite{sakaguchi-etal-2021-proscript-partially, lyu-etal-2021-goal}.
However, because \llm s generate texts in an autoregressive manner based on the previous context, they cannot edit those texts like humans would.
This means that they need to rely on re-generation, which leads to unsatisfactory performance on our proposed task.

\subsection{Task Formulation}
\label{subsec:task}

\begin{figure}[!tbh]
    \centering
    \includegraphics[width=\columnwidth]{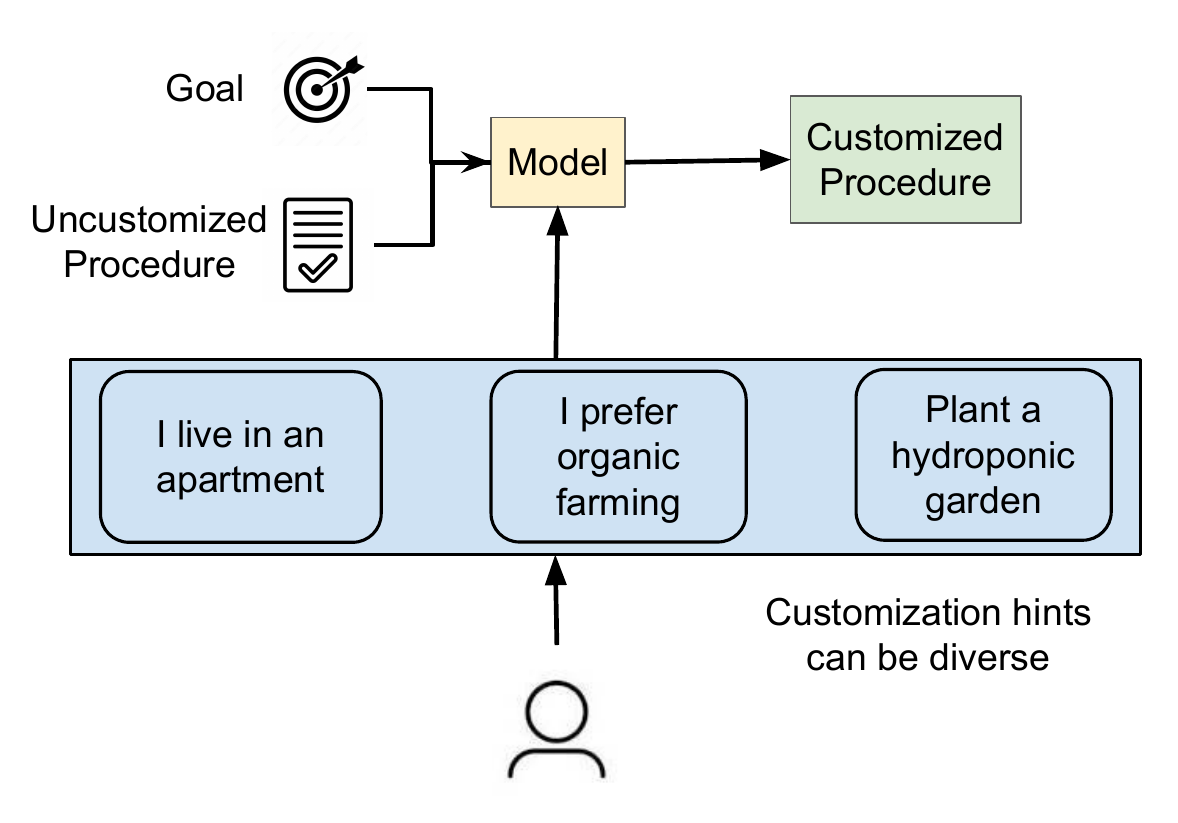}
    \caption{Given a goal $G$, uncustomized procedure $P$ and a user's customization hint $H$, we generate $P'$, a customized procedure that accounts for $H$.}
    \label{fig:task}
\end{figure}

\begin{table*}[!t]
\small
\begin{tabular}
{p{0.10\textwidth}|p{0.10\textwidth}p{0.24\textwidth}p{0.18\textwidth}p{0.25\textwidth}}
\toprule
     Error Type & Goal $G$ & Uncustomized Procedure $P$ & Customization Hint $H$ & Customized Procedure $P'$ \\\midrule

Wrong Order & Bake Bread & 
    1. Dissolve yeast in water.\newline 
    2. Mix the ingredients.\newline 
    3. Knead the dough.\newline
    4. Put dough into greased pan
    \newline ... & I would like to make something sweet \newline and I also like nuts. & 
    ...\newline
    \textcolor{modelred}{3. Place into a baking pan.}\newline
    \textcolor{modelred}{4. Add chopped nuts.}\newline
    \textcolor{modelred}{5. Knead the dough.}\newline ...
 \\\midrule

Extra \newline Steps & Customize Shoes & 
    1. doodle on shoes. \newline
    2. add embellishments. \newline
    3. change out laces for ribbon. \newline
    4. glue rhinestones on straps. \newline
    5. wrap ribbon around straps. & I am a ballet dancer and would like to improve the comfort of my shoes. & 
    1. Identify areas of discomfort.\newline
    2. Purchase gel pads. \newline
    3. Insert in areas of discomfort. \newline
    ... \newline
    \textcolor{modelred}{11. Break in shoes gradually.}
 \\\bottomrule
\end{tabular}
\caption{Example of two major types of errors (shown in \textcolor{modelred}{red}) produced by vanilla \llm s. In the first example (Wrong Order), placing the dough in a baking pan in Step 3 implies that the dough is ready to be baked, so Step 5 (knead the dough) is out of order. In the second example (Extra Steps), there are six extra steps being added (most of them are unnecessary) because the resulting procedure contains 11 steps while the uncustomized one only has five.
}
\label{tab:qual_errors}
\end{table*}

The task, illustrated in \autoref{fig:task}, involves taking as input the goal $G$ of the procedure $P$ as well as a user customization hint $H$.
The output should be a procedure to fulfill the  goal while also taking into account the user scenario described in the hint $P'$.

\subsection{How good are vanilla \llm s on this task?}
\label{subsec:e2e-llms}

We conduct an initial study using vanilla \llm s (see \autoref{fig:task_ex} (left)) on 40 procedures \footnote{These procedures are taken from the new evaluation set \ourdata that we include in this paper (see \S\ref{subsec:data}).}.
Given the goal $G$, uncustomized procedure $P$ and customization hint $H$, the LLM (\turbo) is expected to make changes as per the customization hint.
While this approach is simple, the output is often undesirable (contains extra steps and yet results inadequate or erroneous customization).
\autoref{tab:qual_errors} presents two examples of these errors.
We find that 32.5\% of these errors are due to addition of extra steps which do not apply to $G$, and 15\% of the errors are due to inadequate  or erroneous customization.
Given the shortcomings of end-to-end methods, we need a more structured approach to edit procedures for customization.

\section{Models}

In this section, we describe our multi-agent approach for customizing procedures for users' needs using semi-symbolic edits.
We disentangle the task of generating customized procedures into two aspects, customizability and executability.

\subsection{Agents for Procedure Customization}

Recently, model-based agents \cite{xi2023rise} have been used to perform different types of reasoning in service of achieving a larger goal \cite{yoran-etal-2023-answering}.
We use instances of \llm s to modify for customization and verify for execution and use them in conjunction.
\textbf{\custagent} suggests edits that address a user's customization needs, while \textbf{\execagent} suggests edits to maintain the executability of procedures.
Next, we describe how these agents can interact with each other to best generate customized procedures.

Each agent produces a bag of semi-structured edits $E$ that indicate the operations to be performed, the step in the original procedure $P$ to anchor the edit, and the updated text for the step.
We only allow for two types of edits, insert and replace:
\squishlist
    \item \textit{insert(2, XX)} - a new step with text XX should be added after step 2 of the input procedure.
    \item \textit{replace(3, YY)} - the text of step 3 of the input procedure should be replaced by YY.
\squishend
Note that the replace operation can also perform step deletion by specifying an empty string as its second argument.
The semi-symbolic nature of the edits allows algorithmic application of the edits and decreases model hallucinations compared to end-to-end approaches which make unstructured edits.
We then apply these edits deterministically on $P$ to obtain $P'$ i.e., we find the step number in $P$ that corresponds to a suggested edit and insert/replace it with the edited text.
Having a separate module to apply edits allows us to study the reasoning a model performs when trying to address user requirements.

\subsection{How Do The Agents Interact?}

\begin{figure}[!h]
  \centering
  \includegraphics[width=\columnwidth,height=0.8\columnwidth, keepaspectratio]{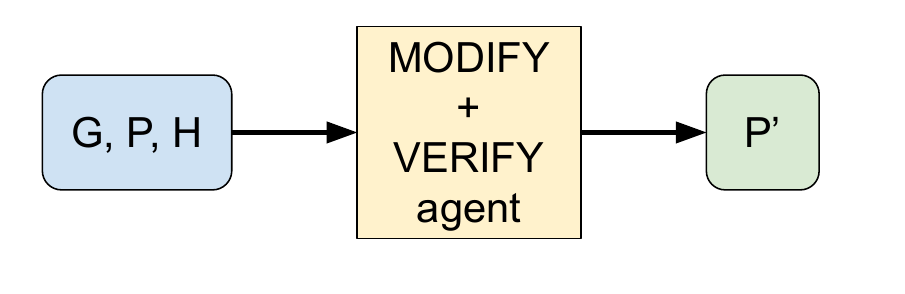}
  \caption{\custagent and \execagent in \uniagent setting. Here, one LLM agent is asked to suggest edits for both customizability as well as executability.}
  \label{fig:unified} 
\end{figure}

\begin{figure}[!h]
  \centering
  \includegraphics[width=\columnwidth,height=0.8\columnwidth, keepaspectratio]{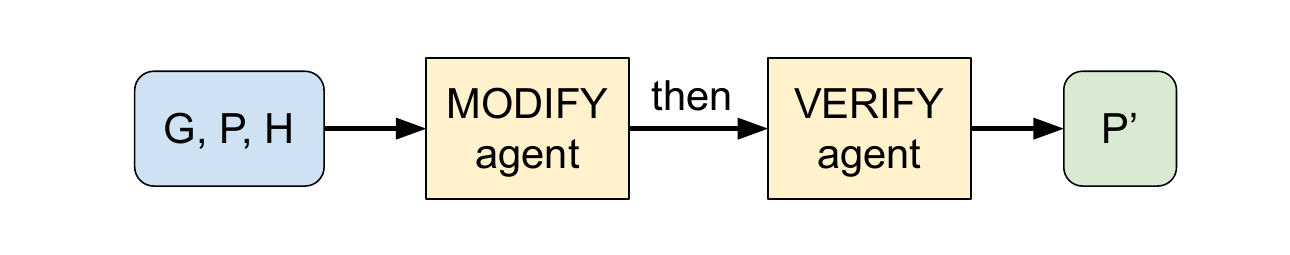}
  \caption{\custagent and \execagent in \seqagents setting. Here, \custagent first generates edits to customize $P$. Then, \execagent makes changes such that the edited procedure is executable, producing $P'$.}
  \label{fig:sequential} 
\end{figure}

\begin{figure}[!h]
  \centering
  \includegraphics[width=\columnwidth, height=0.8\columnwidth, keepaspectratio]{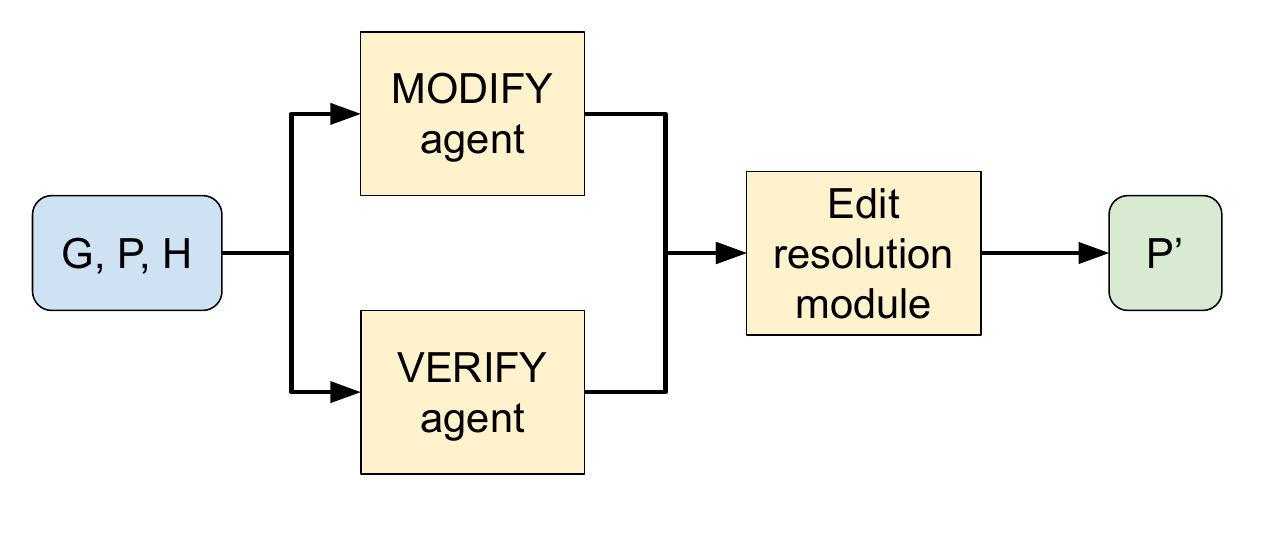}
  \caption{\custagent and \execagent in \paragents setting. Here, \custagent suggests edits to customize the procedure $P$, while \execagent suggests adding often-missing details so that $P$ can be followed. Finally, the edit resolution module takes both sets of edits into account and produces a final bag of edits to be applied to $P$.}
  \label{fig:parallel} 
\end{figure}

We experiment with three ways of interaction for the defined agents: \uniagent, \seqagents, and \paragents, demonstrated in \autoref{fig:unified}, \autoref{fig:sequential} and \autoref{fig:parallel} respectively.

\textbf{\uniagent} - We first define a single agent that is prompted to suggest edits $E$
to $P$ that both customizes it and ensures its executability. 
Mechanically applying these edits results in the customized plan $P'$.
This agent is required to understand how to perform both customization towards a hint $H$ as well as execution to achieve the goal $G$.
This is somewhat similar to the end-to-end method which is also required to understand both aspects of customization. 
However, rather than generating the customized procedure $P'$ directly, it generates
{\it edits} to $P$ that result in $P'$ when applied.
This setting is shown in \autoref{fig:unified}.

\textbf{\seqagents} - In this setting, we first obtain a set of edits $E_c$ from \custagent, and apply $E_c$ to obtain $P_c$.
$E_c$ represents the changes that need to be made in order to suit a user's customization needs.
We obtain $P_c$, a customized procedure, by deterministically applying $E_c$ in $P$.
$P_c$ denotes a customized procedure.
Then, to ensure that this procedure can be executed, \execagent takes $P_c$ as input and suggests another set of edits $E_e$.
Finally, we deterministically apply $E_e$ on $P_c$ to obtain the output customized procedure $P'$.
Here, the agents are used in a sequential order, first suggesting edits to meet customization requirements and next to address any execution-related issues that may arise from those edits.
\autoref{fig:sequential} shows the interaction of the agents in the \seqagents setting.

\textbf{\paragents} - In this setting, shown in \autoref{fig:parallel}, both \custagent and \execagent propose a bag of edits for their respective aspect, $E_c$ and $E_e$, on the uncustomized procedure $P$.
Since changes for customization and execution are different, conflicts arise between those bags of edits.
It is non-trivial to understand how to select the correct edit, since both conflicting edits are important to generate $P'$ but serve different objectives.
To address this, we use an edit resolution module, an LLM that is prompted to take as input two bags of edits, $E_c$ and $E_e$, and produce a merged set of edits $E$. 
Finally, applying $E$ on the uncustomized procedure $P$ results in $P'$.
Functionally, this module is intended to resolve conflicts, merge possible edits and remove redundant edits.
It is also required to remove any edits which cannot be applied to the procedure deterministically.
This is inspired by the meta-reasoner in \citet{yoran-etal-2023-answering}.

\section{Experiments}

To tailor procedures according to customization hints, one needs to make implicit inferences out of the hints, identify the steps that require changes, and finally consistently apply changes to the different steps.
Through our experiments, we aim to understand how well models can modify generic procedures to incorporate aspects captured in customization hints.

\subsection{Our \ourdata evaluation set}
\label{subsec:data}

\begin{figure}[!h]
    \centering
    \includegraphics[width=\columnwidth]{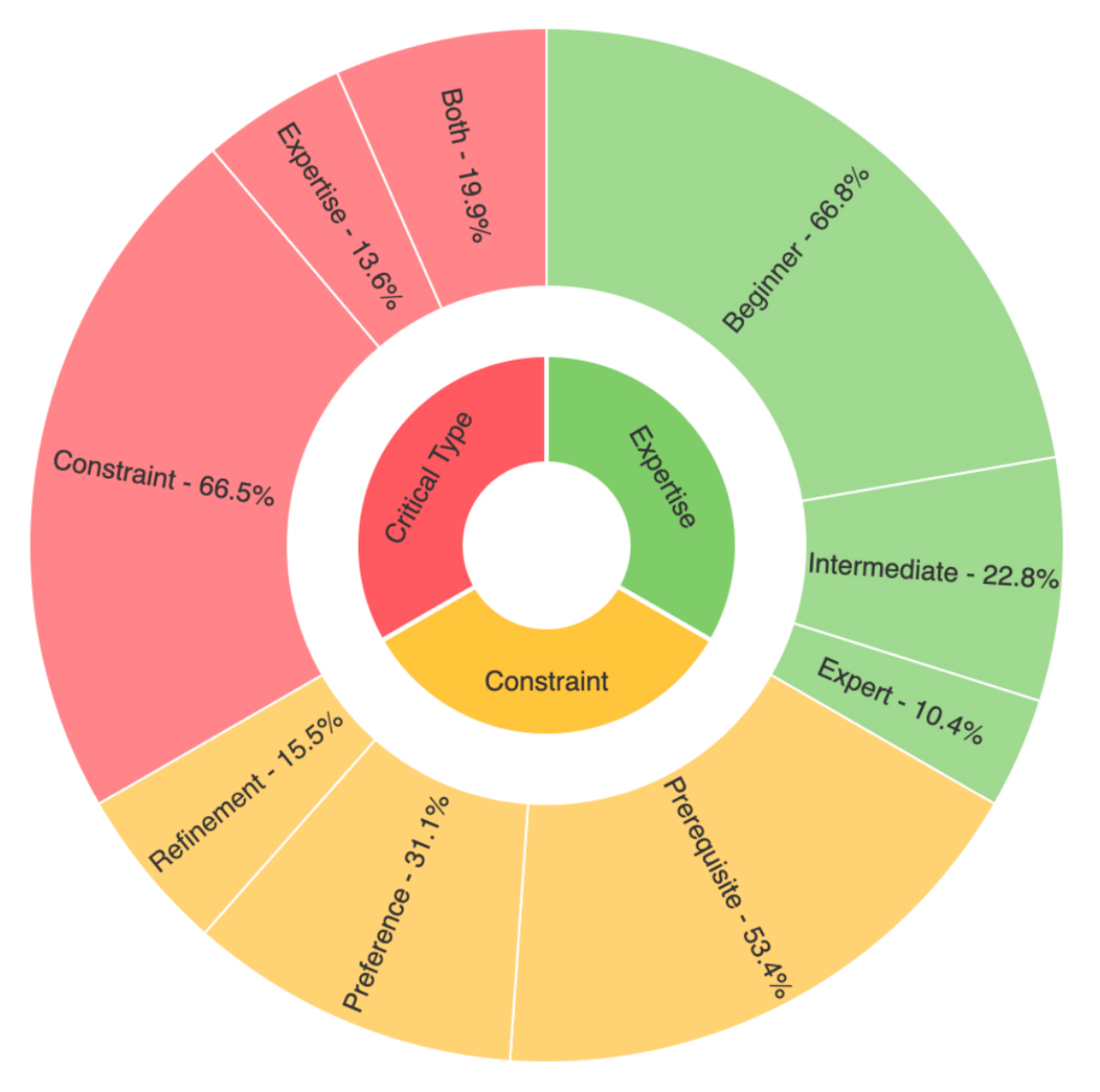}
    \caption{Different types of needs expressed in customization hints in \ourdata.}
    \label{fig:metadata}
\end{figure}

We use WikiHow as the source of diverse goals and corresponding procedures to accomplish them.
Given a broad goal, users are required to write their situation in which they want to accomplish the goal, which acts as their customization hint.
We collect 206 goals over 9 domains, their corresponding WikiHow procedures and customization hints collected from humans to build \ourdata.
Each record is associated with constraint, expertise and critical type (which subtype is more important to perform customization), shown in \autoref{fig:metadata}.
More details can be found in \autoref{sec:dataset}.

\begin{table*}[!t]
    \centering
    \begin{tabular}{|c|r|r|r|}
    \hline
        Method & \cust & \exec & \fc \\ \hline
        E2E Customize & 55.05\% & 48.45\% & 41.46\% \\ \hline
        \seqagents & \textbf{60.68\%} & \textbf{72.33\%} & \textbf{51.94\%} \\
        \uniagent & 54.85\% & 71.36\% & 47.09\% \\ 
        \paragents & 53.88\% & 70.87\% & 45.63\% \\ \hline
        Reverse \seqagents & 42.23\% & 63.59\% & 34.47\% \\ \hline
    \end{tabular}
    \caption{Customizability, executability and fully correct (strictest measure) of procedures generated by different approaches as judged by majority of human evaluators. We note that the \seqagents setting performs the best across all criteria. Note that all approaches built on edit-based agents lead to more executable procedures.}
    \label{tab:main_table}
\end{table*}

\subsection{Evaluation}
\label{subsec:eval}

For open-ended text generation tasks (without gold references) like procedure customization, the absence of an automatic evaluation that correlates well with human judgments is a major challenge \cite{chen-etal-2019-evaluating, ma-etal-2019-results, caglayan-etal-2020-curious, howcroft-etal-2020-twenty, lal-etal-2021-tellmewhy}. 
As a result, we use human evaluation directly, rather than rely on proxies such as \gptfour.

\begin{figure}[!h]
    \centering
    \includegraphics[width=\columnwidth]{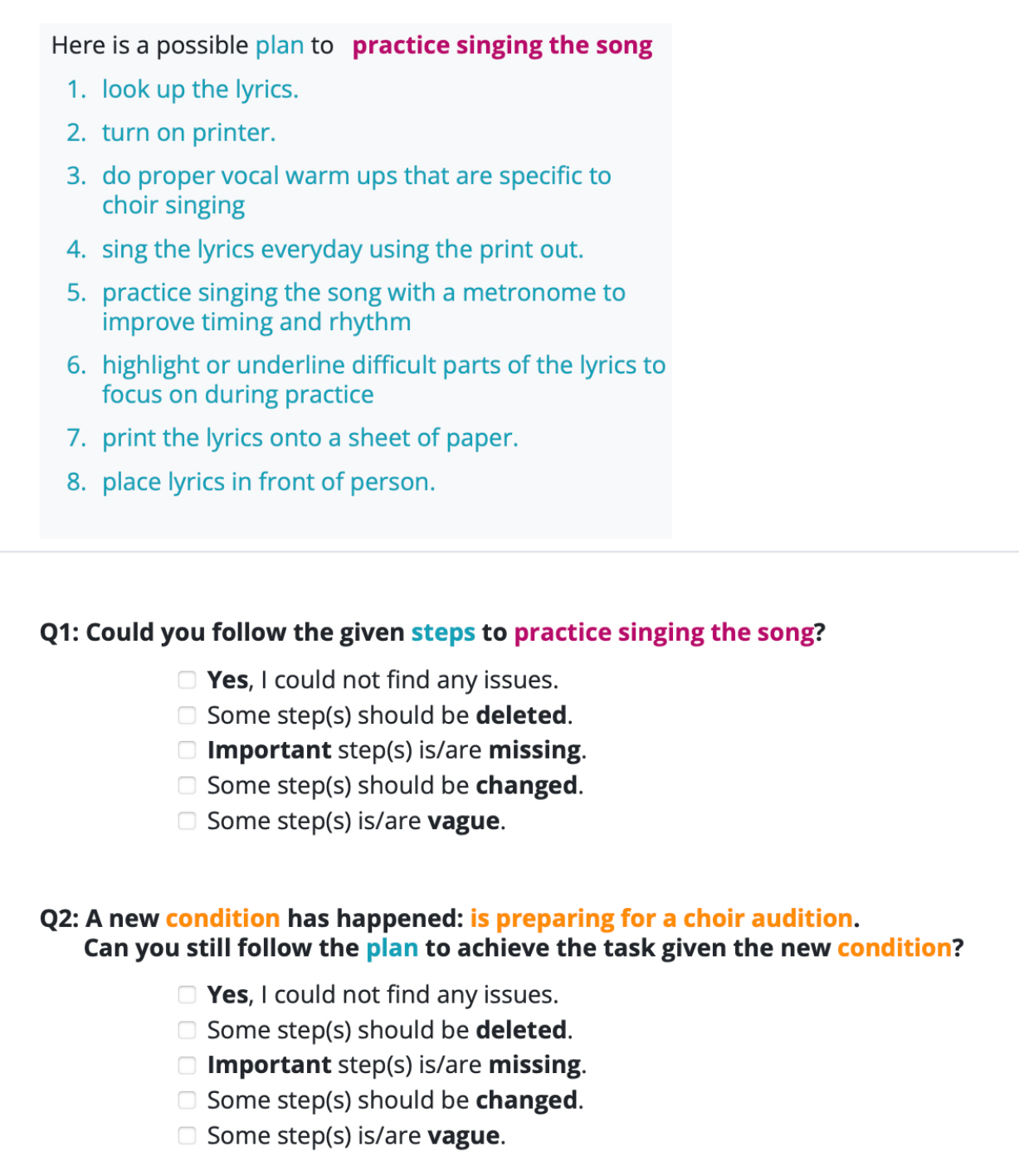}
    \caption{MTurk interface presented to crowdsource workers to judge model generated procedures.}
    \label{fig:crowdsourcing-interface}
\end{figure}

We conduct a human evaluation with a standardized interface to compare different models. 
To this end, we pose questions about customizability, whether a procedure can be performed as is to accomplish the given goal (\exec), and executability, whether it satisfies all the requirements in a presented customization hint (\cust). 
This serves as the human evaluation interface for our task.
The study is illustrated in \autoref{fig:crowdsourcing-interface}.
The task instructions and annotator details can be found in \autoref{sec:mturk_inst}.

For each goal, customization hint, and a model's generated procedure, we ask 3 distinct annotators to provide judgments about customizability and executability of that procedure.
A procedure can be complete in an aspect (customizability or executability), missing steps, have extra steps, have underspecified steps, or have incorrect steps.
An annotator can point out multiple errors about an aspect in the presented plan (negative), or judge the plan to be correct (positive) in that aspect.
We take the majority vote of annotator judgments for the \cust and \exec criteria.
If a plan is judged to be both \cust and \exec by a majority of annotators, we consider it to be \fc.

\subsection{Results}

We report majority statistics for customizability, executability, and correctness of different models in \autoref{tab:main_table}.
Simply using \llm s in an end-to-end fashion is not good enough to generate customized procedures, only generating 41.46\% fully correct procedures. 
We make the following observations. \\

\noindent \textbf{Customize first, fix later}. Using \custagent and \execagent in \seqagents order is the best at producing customized procedures $P'$, generating fully correct customized procedures 51.94\% of the time.
Customizing first allows for making changes to suit a user's customization requirements, before editing the modified procedure to make sure it is executable.

\noindent \textbf{Modifying and verifying together is hard}. Combining both agents into one to obtain a bag of edits in the \uniagent setting requires suggesting edits that serve the purpose of both customization as well as execution.
This is an inherently harder task and, as expected, does not perform as well as the \seqagents setting.

\noindent \textbf{Edit-based customization is interpretable}. Using \llm s as end-to-end customizer is directly comparable to the \uniagent setting.
The former uses natural language while the latter relies on semi-symbolic edits.
By construction, not only does the \uniagent setting perform better, it is also more interpretable since the end-to-end approach sometimes tends to completely change the procedure.

\noindent \textbf{Resolving conflicts is difficult}. 
We find that the \paragents setting is the worst, even though its performance on \cust and \exec are similar to other methods.
However, the intersection of both (\fc) is significantly lower than others.
We hypothesize that this problems arises because the edit resolution module is unable to correctly merge bags of edits from \custagent and \execagent.
This agent can be improved by providing a more complete outlook of how to resolve conflicts and merge relevant edits.

Despite operating on gold procedures, \execagent removes redundant steps or adds critical details to underspecified procedures (often the case with WikiHow).
For example, the instructions to make microwave banana bread start with dry ingredients being mixed in one bowl, while wet ingredients are mixed in another.
However, it does not specify what the dry or wet ingredients are.
The \execagent expands on these details so that the procedure can be followed.
More generally, this framing allows fixing issues in the uncustomized procedure if it isn't from a gold source.

\section{Analysis}

We analyze different aspects of the \seqagents setting and its generated plans.

\begin{figure}[!h]
  \centering
  \includegraphics[width=\columnwidth, height=0.8\columnwidth, keepaspectratio]{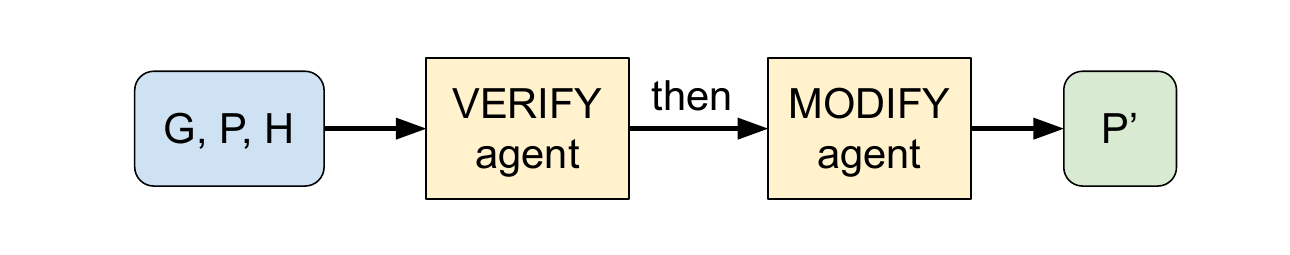}
  \caption{Reverse order of interaction of \execagent and \custagent in \seqagents setting. First, \execagent adds missing details to improve executability, before \custagent proposes changes to suit a user's customization needs.}
  \label{fig:rev_sequential} 
\end{figure}

\noindent \textbf{Ordering of agents matter}. In the \seqagents setting, we flip the order of the \custagent and \execagent, as illustrated in \autoref{fig:rev_sequential}.
We find that this approach only generates usable customized procedures 34.47\% of the time.
While unintuitive, this setting addresses the lack of detail in some WikiHow procedures while also correctly making direct changes for customization.\\

\noindent \textbf{Edits help executability}. 
When comparing the end-to-end approach with any of the edit-based agents, we find that using the agentive framework is better for executability.
\autoref{tab:main_table} shows that all of our approaches produce executable procedures \textgreater 70\% while using \llm s naively generates \textless 50\% procedures that can be followed.

\begin{figure}[!h]
  \centering
  \includegraphics[width=\columnwidth]{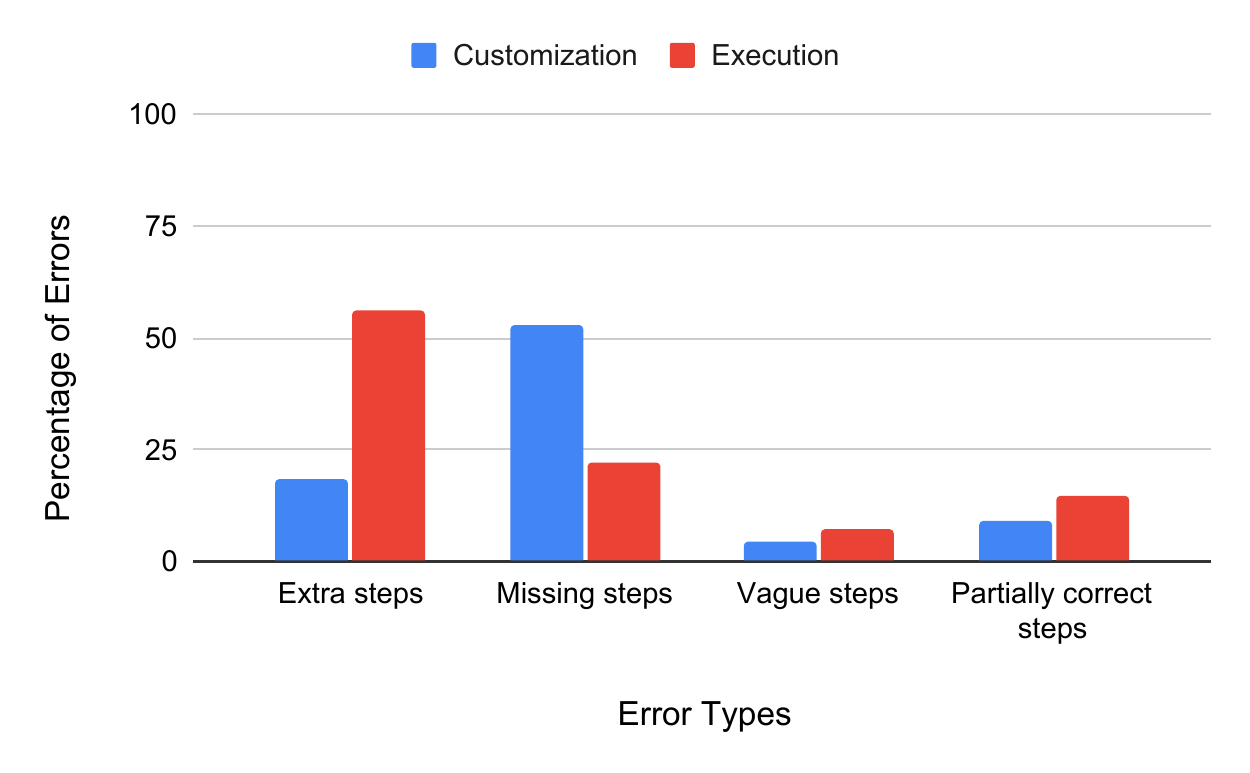}
  \caption{Error distribution in customized plans produced by \seqagents setting.}
  \label{fig:error-dist-sequential} 
\end{figure}

\noindent \textbf{Procedures cannot always be executed}.
We note that execution accuracy for all models is similar.
\autoref{fig:error-dist-sequential} shows that each proposed interaction of the agents suffers the most from generating unnecessary steps, which often hinder achieving the goal of the procedure $G$.

\noindent \textbf{Procedures are not sufficiently customized}.
All methods suffer from the problem of missing steps, indicating that none of them adequately address the requirements stated implicitly or explicitly in the customization hint for the procedure.
We use the \seqagents setting as an illustration in \autoref{fig:error-dist-sequential}.

\subsection{Analyzing subtypes of customization needs}

\ourdata is annotated with different metadata related to types of expertise and constraints.
We use that to perform fine-grained analysis of the generated procedures.

\begin{figure}[!h]
  \centering
  \includegraphics[width=\columnwidth]{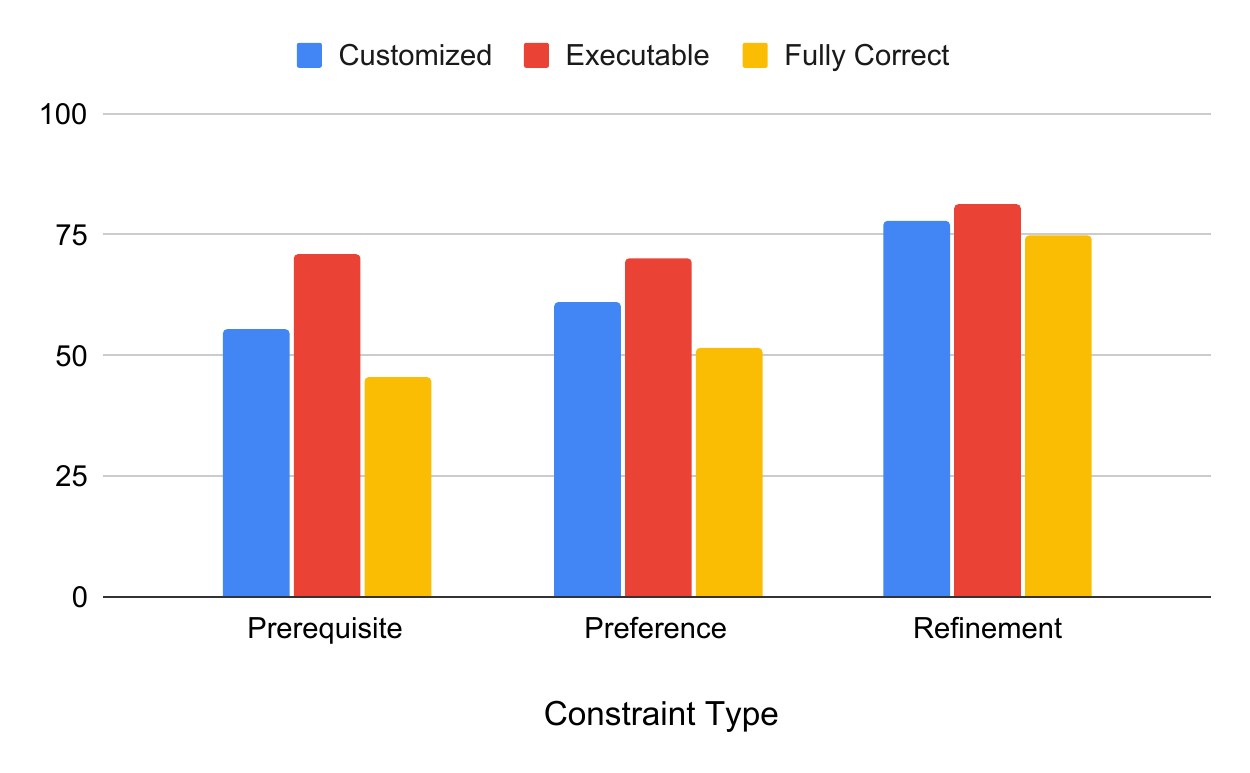}
  \caption{Performance in the \seqagents setting for subtypes of constraints in customization hints.}
  \label{fig:constraint} 
\end{figure}

\noindent \textbf{It is the hardest to satisfy prerequisites}. \autoref{fig:constraint} shows the performance of using the \seqagents setting to address different types of constraints expressed in the customization hint.
Generating customized procedures that address prerequisites is the hardest.
Performance on procedures incorporating preferences is similar.
Upon closer inspection, we see that while the generated procedures might be executable, they do not adhere to the hard constraints set by prerequisities such as an allergy to gluten.

\begin{figure}[!h]
  \centering
  \includegraphics[width=\columnwidth]{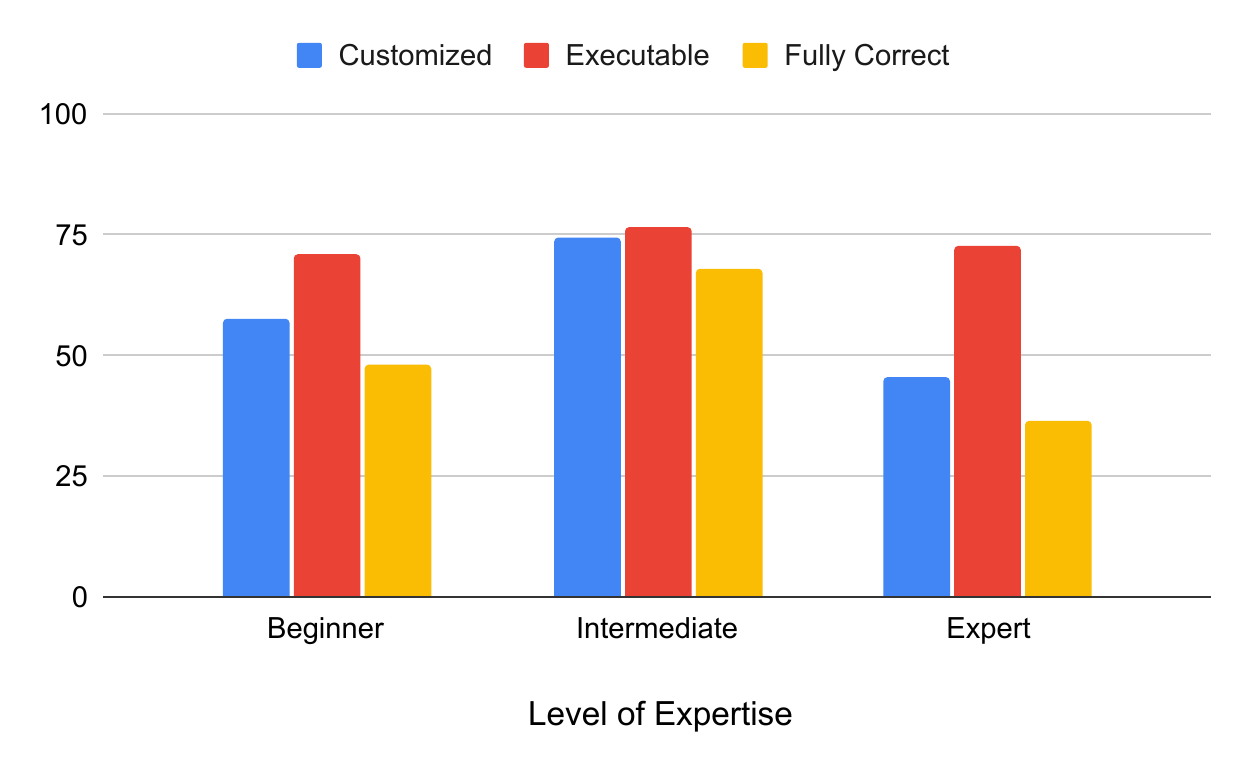}
  \caption{Performance in the \seqagents setting for subtypes of expertise in customization hints.}
  \label{fig:expertise} 
\end{figure}

\noindent \textbf{Complexity of customization matters.}
To study the effect of depth of customization, we compare performance on procedures that require varying degrees of expertise.
It is harder to generate procedures for either experts or for beginners, as presented in \autoref{fig:expertise}.
We hypothesize that domain experts rarely require detailed feedback and can work with small amounts of information to achieve their task.
Conversely, beginners require careful instructions to be able to achieve a goal.
It is difficult for one approach to generate both detailed as well as succinct, high-level plans. \\

\begin{figure}[!h]
  \centering
  \includegraphics[width=\columnwidth]{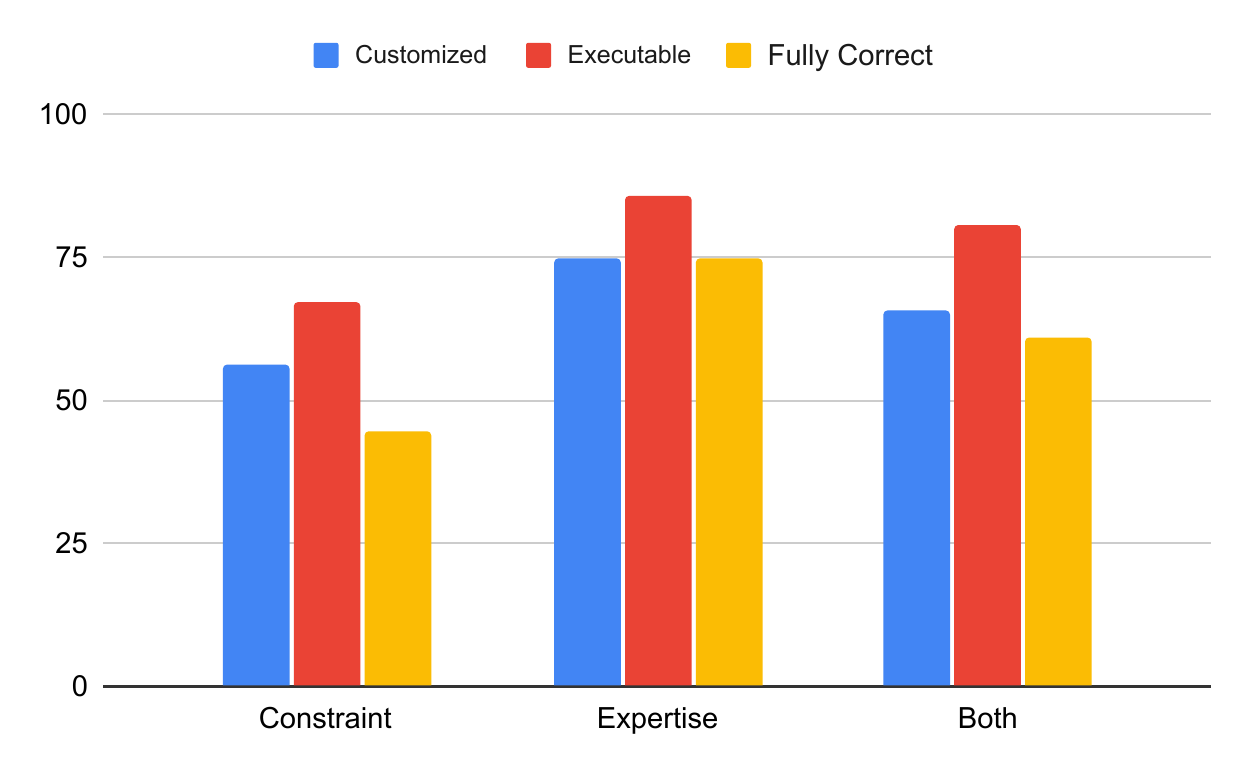}
  \caption{Performance in the \seqagents setting when there are multiple aspects of customization expressed in a customization hint. Among these, constraints are hardest to fully satisfy in the resulting customized procedures.
  }
  \label{fig:critical-type} 
\end{figure}

\noindent \textbf{Constraints are harder to account for than expertise}.
\autoref{fig:critical-type} shows that it is the most difficult to customize according to constraints.
Correctly satisfying requirements expressed in compositional hints, where it is important to consider both expertise and constraint, is also difficult even for the best agent setting.

\subsection{Qualitative Analysis}

On the subset of data points in \S\ref{sec:e2e-llms-eval}, we make binary judgments about customizability and executability, akin to \S\ref{subsec:eval}, after anonymizing the source of the plans to mitigate bias.
The trends for \fc are similar to full-scale evaluation, except that the \paragents setting is slightly better than the \uniagent one.
Executability of procedures on this sample are higher than on the full evaluation set, but trends across methods remain the same, implying that our observations apply to the full \ourdata evaluation set.
We then perform qualitative analysis on plans generated by each approach to understand their characteristics. \\

\noindent \textbf{Edit-based agents are conservative}. 
Edit-based agents tend to suggest less changes than their end-to-end counterparts.
We hypothesize that these agents only generate edits that have higher confidence, and more likely to maintain the coherency of the plan.
On the flip side, it makes it harder to use these agents when the plan needs to be completely changed in order to fully satisfy the requisite customization needs. \\

\noindent \textbf{End-to-end methods have greater creativity}.
While completely rewriting procedures to suit customization needs is not ideal, we observe that it is important to do so when the fundamental way of achieving the goal needs to be altered.
We quantify procedures that require changes in \textgreater 4 steps to fall into this category.
For such procedures, we observe that using \llm s as end-to-end customizers is better since they produce more creative changes while edit-based agents are more conservative.

\subsection{Discussion}

While our task formulation works on uncustomized gold procedures, our edit-based agents can also be used to fix incorrectly customized procedures, since it treats customization as a modification problem.
Instead of using a gold plan as input, our proposed framework supports starting with any related procedure for customization.

More complex tasks would require collaboration between more agents, each dedicated to one aspect of the task.
Despite the effectiveness of the \seqagents setting for procedure customization, it suffers from the problem of agent ordering.
With a larger number of agents, it would be non-trivial to determine the correct order of using agents.

The \paragents setting is flexible and allows the opportunity to integrate edits from multiple agents, similar to prior work \cite{li-etal-2023-making, yoran-etal-2023-answering, wang2023selfconsistency}.
For instance, we can integrate different agents to verify the executability of procedures.
An agent that use entity state changes can be used as a proxy for checking the coherence of a procedure.
Similarly, an agent that enforces the cause and effect relationship between subsequent steps can also be used as a way to verify executability of procedures.
This setting can also be used to integrate different agents for broader applications such as collaborative tool use.
Using edits from such diverse agents is only practically possible when using them in a parallel setting.

\section{Related Work}

Over the past few years, as information about users has been readily available, user-facing technology has become increasingly customized.
Customization has been widely studied in the context of such technology, like search engines \cite{rashid-etal-2002-new-user-pref-ir}, chatbots \cite{majumder-etal-2020-like} and game content \cite{shaker-etal-2010-game-personalized-content}.
There is some work on customized procedural content generation \cite{togelius-etal-2011-personalise-procedures-using-search, yannakakis-etal-2011-experience-based}.
However, they focus on video game content rather than everyday procedures related to pragmatic goals.
Previous research has shown the value of customization in various settings.
\citet{Kapusta2019PersonalizedCP} present a method to customize collaborative plans for robot-assisted dressing.
Building customized care plans is crucial for patients with severe illnesses \cite{Lin2017BuildingPT,Anbari2020BreastCL}.

Procedural text understanding addresses the task of tracking entity states throughout the text \cite{Bosselut2018SimulatingAD, henaff2017tracking}.
Generating such texts \cite{aouladomar-saint-dizier-2005-towards} involves different types of understanding such as temporal reasoning and entity linking.
\citet{mori-etal-2014-flowgraph2text} generated procedures from a meaning representation taking intermediate states into account.
ChattyChef \cite{chattychef} uses the conversational setting to generate cooking instructions and iteratively refine its step ordering.
To study decision branching in procedures, \cite{choice75} generated user scenarios and choices for various steps in a procedure and presented \choice.
For the task of counterfactual planning, \coplan \cite{plasma} collects conditions and combines them with a revised list of steps. 
\citet{majumder-etal-2019-generating} use historical user preferences to generate customized recipes.
But they only focus on one domain (cooking) and on one dimension to model customization.
\llm s have been shown to generate procedural text well but it is unclear how they can be customized for diverse user preferences.

Customized models are designed to capture language patters specific to individual users.
\citet{king-cook-2020-evaluating} examined methods for creating customized LMs using interpolating, fine-tuning and priming.
Similarly, \citet{welch-etal-2022-leveraging} present another approach for fine-tuning and interpolation to customize LMs.
\llm s have also been used to study constrained planning \cite{yuan-etal-2023-distilling} and new interfaces for personalization \cite{ma2023chatbots}.
Role-based prompting is a recent trend.
Pseudo Dialog Prompting \cite{han-etal-2022-meet} is a method to build prompts for LLMs so that chatbots mimic fictional characters, while \citet{vincent-etal-2023-personalised} release a dataset of character annotations to induce personas in LLM output.
However, it is still unclear how to encapsulate more fine-grained aspects of customization.

\section{Conclusion}

This paper studies the capabilities of current \llm s to customize open-domain procedures. 
First , we show that current \llm s cannot effectively customize open-domain procedures.
Next, we propose several \llm-based agent architectures, and find that sequentially using semi-symbolic edits from the \custagent and the \execagent provides an improvement of 10.5\% in generating fully correct procedures over naive prompting.
Even though we show that using edits helps executability, we find that generated procedures are often not sufficiently customized, and there is clear room for improvement.
Finally, we discuss the generalizability of our framework for other diverse applications such as coding and creative writing that require customization. 

\section*{Limitations}

In this work, we focus on using \llm s in a zero-shot setting.
It has been shown that model performance improves by providing in-context examples, performing chain-of-thought reasoning, and incorporating notions of self-consistency.
However, we leave these explorations for future work.

We acknowledge that there can be multiple ways to interpret a customization hint and only the user providing that hint can truly know their needs from the procedure.
This also means that only that person is the right person to evaluate a customized procedure.
Even so, due to the open-domain nature of this task, it is difficult for any one person to evaluate the various aspects of generated procedures.
This can only be done by domain experts.
To alleviate these problems, we use 3 annotators (master turkers) to judge model generated plans and consider their majority judgment \cite{lal-etal-2022-using}, but recognize that this might not be the perfect solution.

Due to the nature of LLMs, they are unable to encapsulate personal preferences \cite{v-ganesan-etal-2023-systematic, dey-etal-2024-socialite}.
Since the procedure customization inherently involves user preferences, LLMs cannot be reliably used as automatic judges yet.

\section*{Ethics Statement}

Our setting assumes that users will not provide any adversarial customization hints.
However, in real-world environments, this assumption is unlikely to hold.
Users can generate malicious procedures by providing such hints to \llm s.

\llm s can generate text that might be biased and insensitive to a user's socio-cultural context \cite{bordia-bowman-2019-identifying, sharma2021evaluating,hovy-prabhumoye-2021-five}.
Since customization hints can have multiple interpretations, it is possible that \llm s can misinterpret the customization needs of users and generate biased and stereotypical procedures.
Thus the system will need checks and balances to ensure that the generated customized procedure is not harmful.

\bibliography{abbreviated_anthology,custom}

\clearpage

\appendix

\begin{figure*}[!t]
    \centering
    \includegraphics[width=\textwidth]{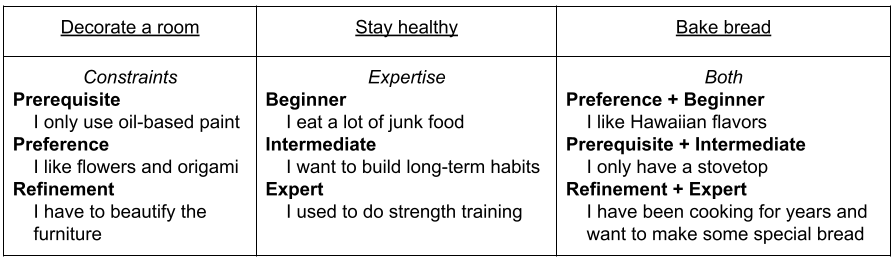}
    \caption{Types of customization hints present in \ourdata, enabling analysis of diverse customization hints. Prerequisites are usually expressed explicitly, while preferences and refinements may be expressed implicitly in the hint. Expertise usually needs to be inferred from the hint, except in hints collected experts. Some hints also encode both user expertise as well as constraints in their scenarios.}
    \label{fig:hierarchy}
\end{figure*}

\section{The \ourdata Evaluation Set}
\label{sec:dataset}

Prior work on editing and customizing procedures has focused on specific domains.
So, data for the task of procedure customization is limited.
To alleviate this problem, we re-purpose two related datasets and collect more high-quality data from WikiHow to create the \ourdata evaluation set.

\subsection{\diagdata}
\label{subsec:ppro}
 
Each WikiHow article consists of a goal (indicated in the title starting with “How to”) and ordered, richly described steps that lead to that goal. 
It already contains articles that include multiple methods for a task.
Additionally, search queries can be used to retrieve and select articles broadly related to the same goal.
However, user interaction signals like fine-grained querying related to specific versions of procedures are not available publicly.
Such signals can be used to improve user experiences.
Therefore, we create a new evaluation set that contains a procedural goal and an individual user's relevant context.

To compile this data, we start by identifying 9 diverse domains.
For each domain, we employ two methods to curate WikiHow articles.
1) Search Results: We devise relevant, popular and pragmatic goals which have relevant instructions present in WikiHow.
We present each goal to users of relevant persona and ask them to provide their feedback according to their personal preferences.
For example, for the broad goal of staying healthy, a collected user persona was that of a teenager looking to build long-term healthy habits. 
2) Multiple Methods: For the same set of broad goals, we filter out WikiHow articles that contain multiple methods to achieve the same goal.
Given the broad goal, users are required to provide situational feedback corresponding to each method.
Using this approach, we collect 106 customized goals (corresponding to user feedback) over 9 domains and build \diagdata, which contains customization hints collected from humans.

\subsection{\easydata}
\label{subsec:coplasma}

To study decision branching in scripts, \citet{choice75} created \choice, a benchmark of 565 data points which requires selecting the next step in a procedure given descriptive scenarios.
These scenarios are valid only at a step level.
However, they can also be incorporated into the full procedure if treated as constraints.
We treat these scenarios as customization hints.

\coplan is a dataset of machine-generated, human-verified scripts.
We use the test set of 861 data points for the counterfactual plan revision task for our purposes.
\citet{plasma} collect goals from diverse topics and prompt \curie to generate a set of ordered steps as a plan to achieve that goal.
\curie is also used, in a few-shot manner, to generate conditions that can alter these plans.
Human verification of these components leads to the creation of a test set that contains a goal, an uncustomized procedure and a relevant condition.
These conditions describe prerequisites to be fulfilled that require the original procedure to be customized in a specific manner.
We treat these conditions as customization hints.
The combination of data from these two sources is referred to as \easydata since it contains customization hints generated automatically.

\subsection{Statistics: \ourdata evaluation set}

We randomly select 100 procedures and their corresponding customization hints from \easydata and add them to \diagdata to form the \ourdata evaluation set.
Overall, \ourdata contains 206 data points, each of which is made up of a goal, a list of steps to achieve that goal and a customization hint  from a user according to which the procedure should be modified.
It contains 106 unique goals and 203 unique customization hints.

We also store relevant metadata for each user persona providing the annotation such as their level of expertise and their constraints like dietary preferences or availability of tools.
6.2\% of these customized procedures can only be performed by domain experts, 11.6\% need an intermediate level of expertise, and the rest are beginner-friendly.
The user comments contain implicit or explicit constraints --- 58\% contain hard prerequisites, 30\% reflect some user preference and the rest mention a target refinement to be achieved. 
The types of customization hints, as well as corresponding examples, in this dataset are presented in \autoref{fig:hierarchy}.

\section{Mechanical Turk tasks}
\label{sec:mturk_inst}

\begin{figure*}
    \centering
    \includegraphics[width=\textwidth]{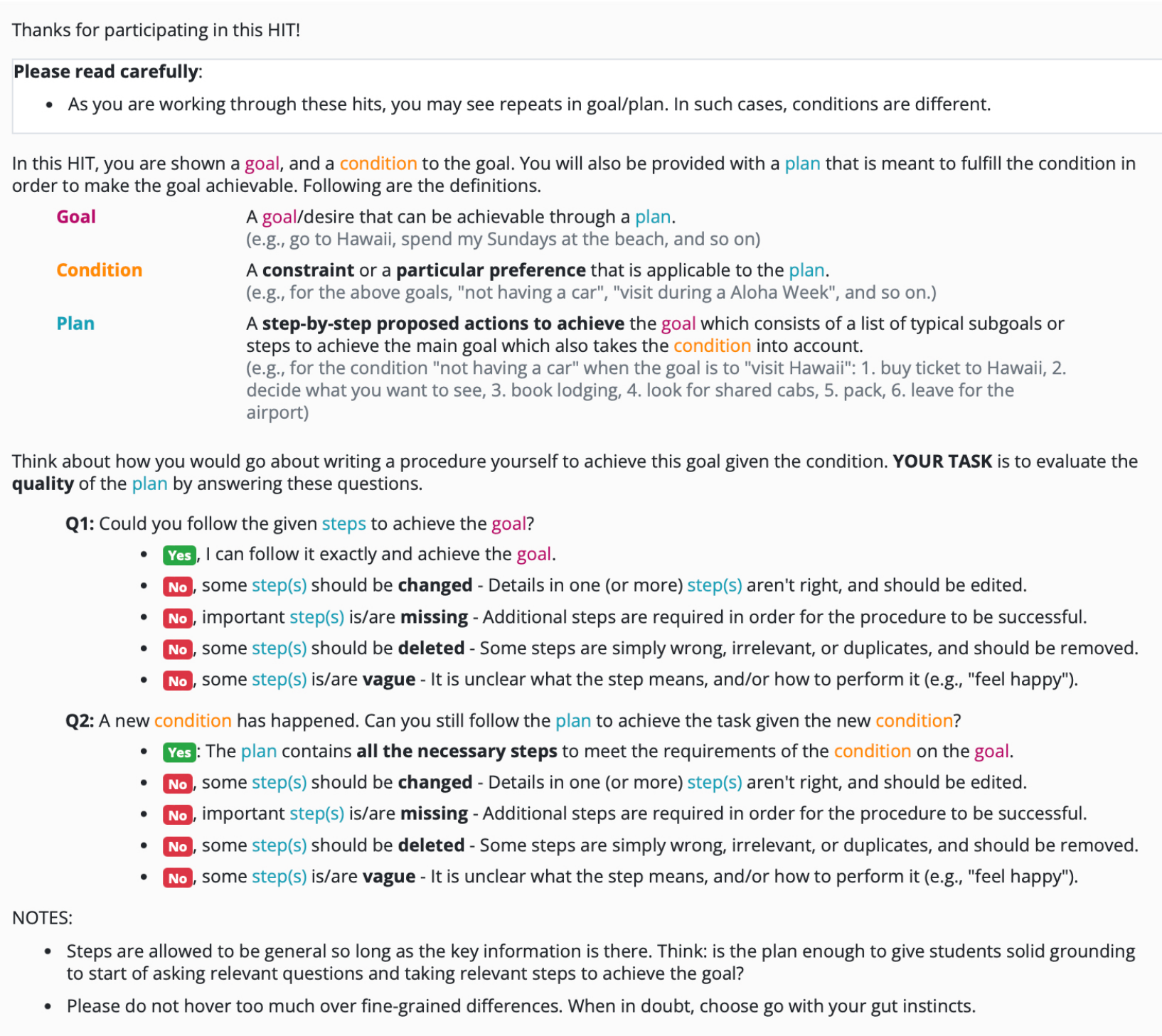}
    \caption{Instructions for MTurk tasks}
    \label{fig:inst}
\end{figure*}

We present the instructions given to annotators for both the tasks in \autoref{fig:inst}.
Annotators were given clear direction for the task, as well as provided examples for various cases of desirable and undesirable characteristics of a procedure.
We restricted the task to master turkers.
The master turkers are paid 0.30\$ per HIT, which translates to 17\$/hr according to average time needed for completion.

\section{Reproducibility}

\subsection{LLM Settings}

We used a temperature of 0.0 for all the experiments to select the most likely token at each step, as this setting allow for reproducibility\footnote{We note that some researchers have shown that even this setting might not make it completely reproducible: \url{https://twitter.com/ofirpress/status/1542610741668093952?s=46&t=f9v5k9RzVKnTK1e0UyauOA} and \url{https://twitter.com/BorisMPower/status/1608522707372740609}}. 

We use the following code snippet for any experiments performed with \turbo:
{\small
\begin{verbatim}
import openai
openai.api_key = os.getenv("OPENAI_API_KEY")
response = openai.ChatCompletion.create(
            model="gpt-3.5-turbo-0301",
            messages=[{'role': 'user', 
                       'content': prompt}],
            temperature=0.0, # reproducibility.
            max_tokens=500, 
            top_p=1, 
            frequency_penalty=0.1,
            presence_penalty=0
            )
\end{verbatim}
}

\subsection{Prompts Used}

\begin{figure*}
    \centering
    \includegraphics[width=\textwidth]{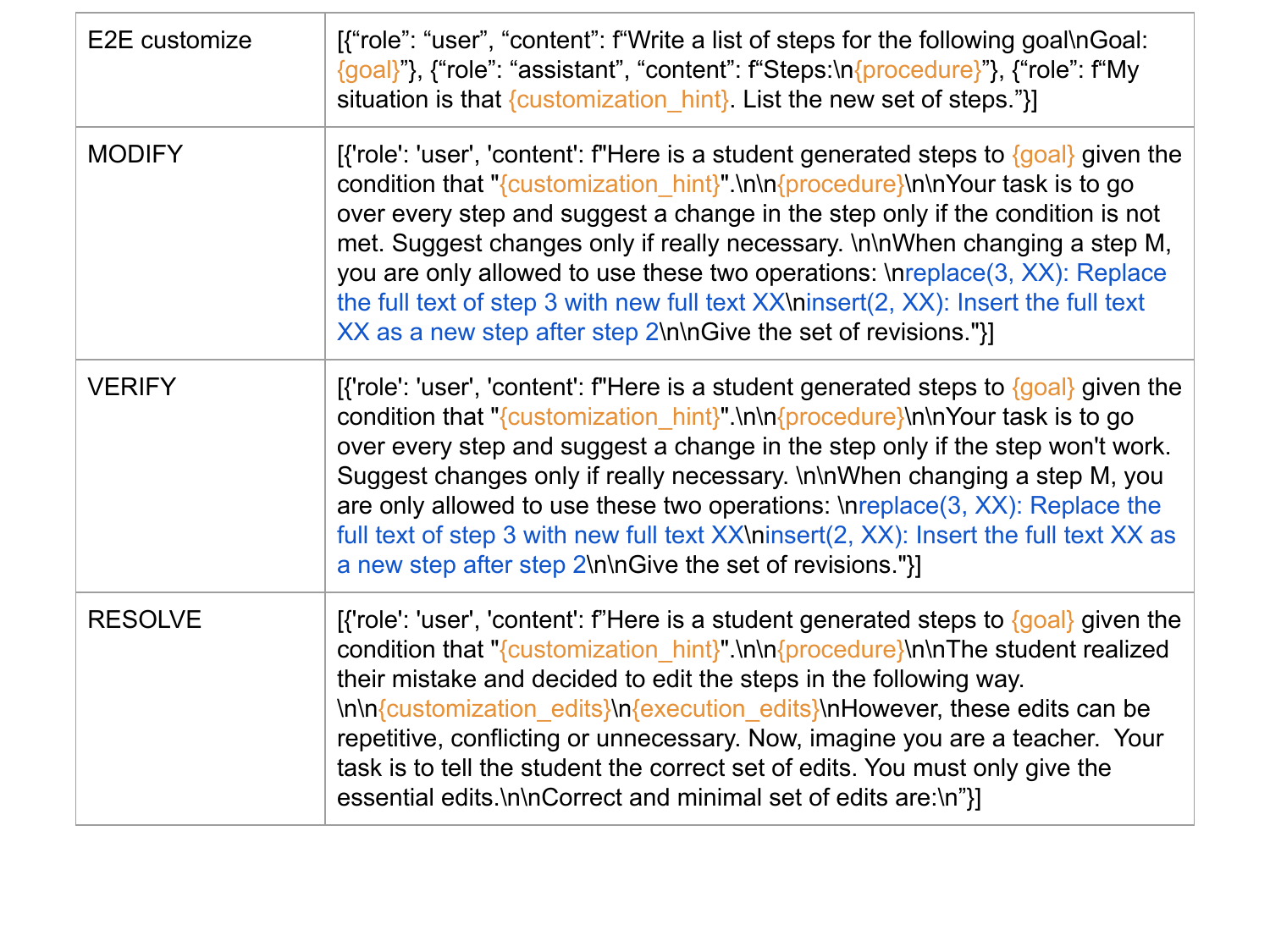}
    \caption{Prompts used in various parts of our experiments.}
    \label{fig:prompts}
\end{figure*}

We list the prompts we use in different parts of our methods in \autoref{fig:prompts}.

\end{document}